\documentclass[a4paper]{article}

\usepackage{INTERSPEECH2019}
\usepackage{multirow}
\usepackage{graphicx}

\title{Probing the phonetic and phonological knowledge of tones in Mandarin TTS models}
\name{Jian Zhu}
\address{
  University of Michigan, Ann Arbor, USA}

\email{lingjzhu@umich.edu}

\begin{document}

\maketitle
\begin{abstract}
This study probes the phonetic and phonological knowledge of lexical tones in TTS models through two experiments. Controlled stimuli for testing tonal coarticulation and tone sandhi in Mandarin were fed into Tacotron 2 and WaveGlow to generate speech samples, which were subject to acoustic analysis and human evaluation. Results show that both baseline Tacotron 2 and Tacotron 2 with BERT embeddings capture the surface tonal coarticulation patterns well but fail to consistently apply the Tone-3 sandhi rule to novel sentences. Incorporating pre-trained BERT embeddings into Tacotron 2 improves the naturalness and prosody performance, and yields better generalization of Tone-3 sandhi rules to novel complex sentences, although the overall accuracy for Tone-3 sandhi was still low. Given that TTS models do capture some linguistic phenomena, it is argued that they can be used to generate and validate certain linguistic hypotheses. On the other hand, it is also suggested that linguistically informed stimuli should be included in the training and the evaluation of TTS models. 
\end{abstract}

\section{Introduction}
Recent advancements in end-to-end text-to-speech (TTS) synthesis (i.e., Tacotron 2 \cite{shen2018natural}, TransformerTTS \cite{li2018close}, FastSpeech \cite{ren2019fastspeech}) and waveform generation (e.g., WaveNet \cite{oord2016wavenet}, SampleRNN \cite{mehri2016samplernn}, WaveGlow \cite{prenger2019waveglow}) have made it possible to generate extremely natural speech that can sometimes even fool human evaluators. Given such impressive performance, these models have necessarily captured at least some, if not all, statistical regularities in the highly varying speech signals. One is therefore naturally inclined to ask: to what extent are these deep neural networks capturing the statistical regularities and/or irregularities of speech? 

While such works are lacking in speech science, researchers in natural language processing (NLP) have investigated the extent to which the behaviors of neural language models are comparable to human syntactic abilities. They evaluate deep language models with controlled psycholinguistic stimuli, showing that at least some aspects of neural networks are similar to human language processing, despite key differences \cite{futrell2018rnns,gulordava2018colorless,prasad2019using}. While these studies focus on whether neural networks learn human-like linguistic abilities, it has been suggested that neural networks can also benefit language research by providing computational models for generating or testing linguistic hypotheses \cite{linzen2019can}. Recent investigations have applied representation learning to study tones and intonations in several languages \cite{li2019representation,gerazov2018variational}.

The current work goes one step further to ask: is it possible to exploit these learned parameters to gain a better understanding of speech? TTS models seem to combine the advantages of both experimental and corpus-based approaches. They are trained on many hours of speech and therefore are potentially more generalizable to diverse linguistic patterns. Once a TTS model is trained, it can be used to generate speech samples from texts unseen in the training data. If the quality of generated samples is good enough, TTS models can be employed to generate speech from controlled sentence stimuli, which can be subjected to the same acoustic analysis and statistical testing as in most phonetic experiments. However, before this approach is applied to linguistic research, it is important to understand what phonetic and phonological knowledge these TTS models have learned.

\section{The current study}
To validate the approach outlined in the introduction, this study assesses TTS models' learned knowledge of tonal coarticulation and Tone-3 sandhi in Mandarin with controlled stimuli. In Mandarin, lexical meanings are distinguished through four contrastive tones (Tone-1: high-level; Tone-2: rising; Tone-3: dipping; Tone-4: falling), though some suggest that a fifth neural tone also exists \cite{chen2006production}. Tonal coarticulation refers to the within-category variation of lexical tones induced by neighboring tones, which will not change the tone type. Tone sandhi is the mandatory categorical change of tone types caused by neighboring tones. In Mandarin, the most common example is Tone-3 sandhi, in which Tone-3 changes to Tone-2 if it precedes another Tone-3. This process is automatic and mandatory, although subject to some syntactic constraints.

Tonal coarticulation and tone sandhi represent gradient and categorical tonal variations in Mandarin respectively. Both have been extensively investigated for Mandarin \cite{xu1997contextual,shen1990tonal,chien2016priming,shih1997mandarin}, making them suitable probes to investigate the phonetic and phonological knowledge of TTS models and verify the proposed approach.

The current work has the following contributions. This is among the first studies to probe the phonetic and phonological knowledge of TTS models, a step towards better understanding of the inner workings of neural networks from a linguistic perspective. It shows that TTS models can be used in linguistic research to explore certain hypotheses because they capture well the gradient phonetic processes in surface form, though still falling short of learning the underlying categorical phonological rules. Secondly, this work also highlights the importance of designing linguistically informed stimuli to train evaluate TTS models.

\section{Methods}
In this study, stimuli are fed into TTS models to generate speech samples, which are later analyzed as in linguistic experiments.  A speech synthesis model (here, Tacotron 2 \cite{shen2018natural}) takes textual stimuli as input to predict the corresponding mel-spectrogram, and then the log mel-spectrogram is converted to raw waveform through a vocoder (here, WaveGlow \cite{prenger2019waveglow}).

\subsection{Stimuli}
\subsubsection{Tonal coarticulation}
The goal of this experiment is to examine both anticipatory and carry-over tonal coarticulation in Mandarin. Tones in Mandarin can be characterized by their onset and offset values (Table \ref{tab:onoff}) \cite{xu1997contextual}. Anticipatory coarticulation refers to the influence of the onset value of the upcoming tone on the target tone, whereas carry-over coarticulation refers to the tonal variation due to the offset value of the previous tone. The stimuli were all possible bisyllabic combinations of syllables [ma, mo, mi] with different tones embedded in 6 carrier phrases, resulting  in 576 sentences in total. Most of these words were designed to be non-sense to exclude the influences of extraneous factors (e.g., frequency, semantic predictability).

\begin{table}[th]
  \caption{Onset and offset of Mandarin tones.}
  \label{tab:onoff}
  \centering
    \begin{tabular}{rlrr}
    \toprule
                       &      & \multicolumn{2}{c}{Offset} \\ \cline{3-4} 
                       
                       &      & High         & Low         \\
    \midrule
    \multirow{2}{*}{Onset} & High &   Tone 1      &    Tone 4      \\
                            & Low  &  Tone 2     &   Tone 3   \\
    \bottomrule
  \end{tabular}
  \label{tab:tones}
\end{table}

\subsubsection{Tone-3 sandhi}
This experiment assesses whether TTS models can correctly perform the Tone-3 sandhi rules without explicit supervision. Let $\sigma$ represent a syllable which can carry different lexical tones ($\sigma_1,\sigma_2,\sigma_3,\sigma_4$) and brackets $[\quad]_w$ represent a word. If two Tone-3 syllables co-occur, the first one will change to Tone-2, that is, $[\sigma_3\sigma_3]_w \rightarrow [\sigma_2\sigma_3]_w$. However, when three Tone-3 syllables co-occur, there are two patterns of tone change depending on the syntactic constituency (left-branching or right-branching). 
\begin{itemize}
    \item $[\sigma_3\sigma_3\sigma_3]_w \rightarrow [\sigma_3[\sigma_2\sigma_3]_w]_w$  \\
    Ex. mi3 lao3 - shu3 'Micky mouse' $\rightarrow$ mi3 lao2 - shu3
    
    \item $[\sigma_3\sigma_3\sigma_3]_w \rightarrow [[\sigma_2\sigma_2]_w\sigma_3]_w$. \\
    Ex. meng3 - gu3 yu3 'Mongolian' $\rightarrow$ meng2 - gu2 yu3
\end{itemize}

In other words, syntactic analysis is required to correctly apply the Tone-3 sandhi rule \cite{shih1997mandarin} in phrases. Challenging and uncertain cases arises when Tone-3 sequences occur in more complex syntactic structures. Individual differences also exist, as $[\sigma_3]_w[\sigma_3]_w[\sigma_3]_w$ can be realized as any one of the above forms because of speaker choice.

The stimuli for Tone-3 sandhi were 38 bisyllabic words, 32 trisyllabic and 39 phrases that range from 4 to 19 characters, all of which were in Tone-3. The list was compiled through gathering stimuli from previous studies \cite{chien2016priming,shih1997mandarin} or online corpora.  

\subsection{Tacotron 2 with BERT embeddings}
The end-to-end speech synthesis model used in this study, Tacotron 2 \cite{shen2018natural}, is a state-of-the-art speech synthesis model. As a sequence-to-sequence recurrent neural network, Tacotron 2 can predict log mel-spectrograms from only raw textual inputs. The alignment between textual inputs and predicted spectrograms are learned through location-sensitive attention mechanism.

\begin{figure}
    \centering
    \includegraphics[width=\linewidth]{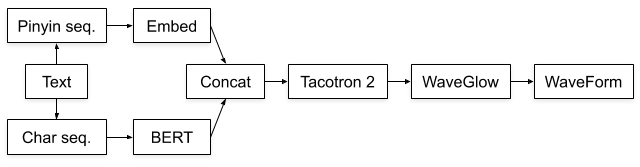}
    \caption{Workflow of Tacotron 2 + BERT and WaveGlow}
    \label{fig:model}
\end{figure}

The original Tacotron 2 only requires phonemes or characters as input. Recent works \cite{fang2019towards,hayashi2019pre,tyagi2019dynamic} have explored using syntactic and semantic representations from Bidirectional Encoder Representations from Transformers (BERT), the state-of-the-art pre-trained language model for textual feature extraction \cite{devlin2018bert}, achieving better quality than phoneme/character-only models. In this experiment, two Tacotron 2 models were trained, Tacotron 2 with only Pinyin inputs and Tacotron 2 with Pinyin and BERT embeddings. Given the large number of characters, it is unrealistic to use character-based features in Mandarin TTS. Pinyin (phonemic) features are instead adopted as the input features. The syllable structure of Mandarin is relatively simple, as each Chinese character is represented by either one or two Pinyin symbols, a combination of initials and finals, or onsets and rhymes. No explicit tone sandhi information was supplemented in the grapheme-to-phoneme process, as TTS models were expected to pick up the tone sandhi rules in a purely data-driven approach. The BERT embeddings are extracted for each character with the character-based Chinese BERT \cite{cui2019pre}. Finally they are upsampled to match the length of the Pinyin symbols of each character. Then the BERT embeddings and phonemic embeddings are concatenated as the input to the subsequent layers (Fig. \ref{fig:model}). Except for the inputs and the first convolutional layer that mapped inputs into lower dimensions, both TTS models have the exact same architecture and hyperparameters.

\subsection{WaveGlow}
WaveGlow is a non-autoregresssive vocoder capable of converting mel-spectrograms to waveforms faster than real time \cite{prenger2019waveglow}. Previous vocoders such as WaveNet \cite{oord2016wavenet} and SampleRNN \cite{mehri2016samplernn} can generate extremely natural speech waveforms from mel-spectrograms but are painfully slow because of the autoregressive generation process. WaveGlow combines the flow-based generative modeling \cite{kingma2018glow} and WaveNet to achieve non-autoregressive waveform generation, making it possible to massively speed up the generation process while preserving the naturalness of synthesized speech. In this study, a WaveGlow vocoder for Mandarin speech was trained on log mel-spectrograms that match the outputs of Tacotron 2.

\section{Experiments}
The code, pre-trained models, stimuli and detailed experimental settings are all available online at [https://github.com/lingjzhu/probing-TTS-models].

\subsection{Data}
The training data for both Tacotron 2 and WaveGlow was the Chinese Standard Mandarin Speech Corpus (CSMSC)\footnote{https://www.data-baker.com/open\_source.html}. CSMSC has 10,000 recorded sentences read by a female speaker, totaling 12 hours of natural speech with phoneme-level Textgrid annotations and text transcriptions. The corpus was randomly partitioned into non-overlapping training, development and test sets with 9800, 100, 100 sentences respectively.

\subsection{Model training}
\subsubsection{Tacotron 2}
The Tacotron 2 was based on the NVIDIA's public implementation and the pre-trained model\footnote{https://github.com/NVIDIA/tacotron2}. The model was trained with a batch size of 64 and the Adam optimizer (lr=1e-3). Weights of the Tacotron 2 model were initialized with the pre-trained English model for compatible layers. It is found that weights for English speech synthesis can accelerate the convergence of the Mandarin model. The model began to form diagonal attention alignment in one thousand iterations and produce intelligible speech after only several thousand iterations. The Mandarin Tacotron 2 was trained for 50k iterations on a single RTX 2080Ti. The entire training took less than 2 days to complete. 

\subsubsection{WaveGlow}
All default settings of the original WaveGlow implementation\footnote{https://github.com/NVIDIA/waveglow} were kept. The batch size was set to 8 and the Adam optimizer with a learning rate of 1e-4 was used. The Mandarin WaveGlow model was also initialized with the weights of the pre-trained English WaveGlow, because transferring weights from a pre-trained model, even in a different language, greatly speeds up the training process. With the warm start, the model produced good-quality speech at only 46k iterations but the final model was trained for 150k iterations. The training process was done on a single RTX 2080Ti GPU, which lasted 3 days.

\begin{figure}
    \centering
    \includegraphics[width=0.45\linewidth]{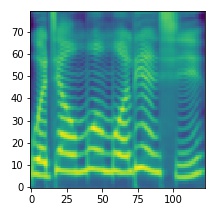}
    \includegraphics[width=0.45\linewidth]{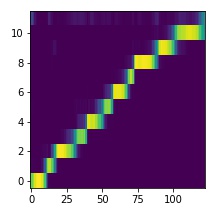}
    \caption{Predicted log mel-spectrogram (left) and attention alignment plot (right) for \textbf{wo3 jiao4 mo4 mo4 lian4 xi2} ("I asked mo mo to practice").}
    \label{fig:pred}
\end{figure}

\subsection{Evaluation}
Acoustic analysis was carried out to examine the generated speech samples for tonal coarticulation. The time-aligned Textgrid annotations for each stimulus were derived from the attention alignment (Fig. \ref{fig:pred}) through a simple greedy approach. Attention alignment does not always translate to accurate transcriptions so the generated Textgrids were inspected. Three sentences were excluded due to problematic annotation. The f0 tracking was done using the auto-correlation method through the Parselmouth API of Praat \cite{jadoul2018introducing}.

The rest of the speech samples were assessed by five trained linguists who were also native speakers of Mandarin Chinese. They were not informed of the purpose of this study.
\subsubsection{MOS evaluation}
The performance of the overall quality samples was evaluated using the mean opinion score (MOS). Listeners were asked to rate the overall naturalness and prosodic appropriateness of samples on a scale from 1 and 5. Speech samples for the CSMSC test set were generated using both Tacotron 2 models. Two sentences in the test set were removed due to potentially inappropriate content. Then these synthesized samples were mixed with real speech samples and presented to listeners independently in random order.
\subsubsection{Tone-3 sandhi}
In this task, listeners were asked to judge whether there were Tone-3 sandhi errors in the speech samples while ignoring any acoustic artifacts not related to lexical tones. Specifically, for phrase level stimuli, they were also asked to identify the number of tonal errors in the samples.

\section{Results}
\subsection{MOS}
Table \ref{tab:mos} summarizes the results of listeners' evaluation of the generative samples. As suggested in \cite{rosenberg2017bias}, paired Mann-Whitney U Tests were used to test pairwise comparisons for naturalness and prosody over three types of speech samples, and all comparisons were significant (p $\leq$ .001) after applying the conservative Bonferroni correction for multiple comparisons. These results show that both models can synthesize highly natural speech samples with acceptable prosody and, similar to previous studies \cite{hayashi2019pre,fang2019towards}, Tacotron 2 with BERT embeddings improved the naturalness and prosody of generated speech over the baseline Tacotron 2. It is also noticed that Tacotron 2 with BERT embeddings tends to make slightly more pronunciation errors than the baseline Tacotron 2.
\begin{table}[th]
  \caption{Mean MOS of speech samples}
  \label{tab:example}
  \centering
  \begin{tabular}{ r@{}r  r }
    \toprule
    \multicolumn{1}{c}{\textbf{Models}} & 
    \multicolumn{1}{c}{\textbf{Naturalness}} &
    \multicolumn{1}{c}{\textbf{Prosody}} \\
    \midrule
    Tacotron 2 & 3.65 & 3.86 \\
    Tacotron 2+BERT & 4.04 & 4.21 \\
    Ground Truth & 4.39 & 4.53 \\
    \bottomrule
  \end{tabular}
  \label{tab:mos}
\end{table}

\subsection{Tonal coarticulation}

\begin{figure}[tbh]
    \centering
    \includegraphics[width=\linewidth]{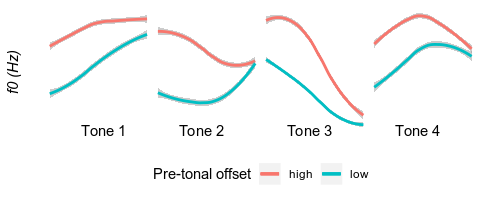}
    \caption{Assimilatory carry-over coarticulation produced by Tacotron 2. Tone onsets are strongly affected by the offsets of previous tones.}
    \label{fig:ass}
\end{figure}

\begin{figure}[tbh]
    \centering
    \includegraphics[width=\linewidth]{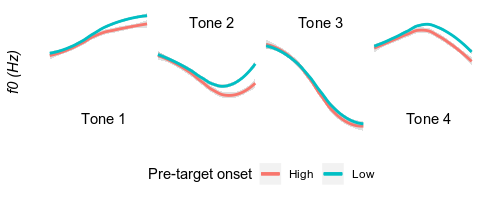}
    \caption{Dissimilatory anticipatory coarticulation produced by Tacotron 2. Tone offsets are only slightly affected by the onsets of upcoming tones.}
    \label{fig:diss}
\end{figure}

As both models capture equally well the surface variations of lexical tones, only results from Tacotron 2 are reported for illustration. Regression models were fitted to each tone in each condition through locally estimated scatterplot smoothing (LOESS) and resulting curves with shaded 95\% confidence intervals were presented in Fig. \ref{fig:ass} and \ref{fig:diss}. There are strong (assimilatory) carry-over effects of the offset of the previous tone (Fig. \ref{fig:ass}), as the overall pitch is significantly lowered or raised by the low or high offset of the previous tone, respectively. Tone-3 + Tone-3 combinations were excluded in the analysis because it is a categorical tone sandhi process examined in next section. Fig. \ref{fig:diss} indicates that anticipatory effects are weaker than the carry-over effects: the tone contours do not differ much before different onsets of the upcoming tone. Moreover, the anticipatory coarticulation is mostly dissimilatory, in which low onset f0 of the upcoming tone raises the f0 maximum of the previous tone. The findings closely mirror those obtained from native speakers \cite{xu1997contextual} even in fine-grained details, suggesting that these coarticulatory patterns are learned and faithfully reproduced in TTS models in novel sentences.

\subsection{Tone-3 sandhi}

The Tone-3 sandhi experiment presents an interesting case concerning the phonological ability of TTS models. Mixed logistic regression models were fitted to test whether the difference between two models was significant. Both TTS models made few Tone-3 sandhi mistakes on bisyllabic words, even when such information is not specified in the Pinyin inputs. But Tacotron 2 is more accurate in synthesizing bisyllabic Tone-3 words than Tacotron 2 with BERT embeddings ($\hat{\beta}=2.35,p\leq0.001$).

The performance decreased for more complex stimuli, in which tone sandhi patterns interact with syntactic structures. Both models perform nearly equally poorly on trisyllabic words ($\hat{\beta}=-0.05,p=0.82$), with mistakes unlikely to be made by native speakers. However, Tacotron 2 with BERT embeddings performs better than the phoneme-only Tacotron 2 on complex sentences ($\hat{\beta}=-0.52,p=0.01$). On average, Tacotron 2 with BERT made 0.92 mistakes per phrase, while the baseline Tacotron 2 had 1.31 mistakes per phrase. This implies that at least some syntactic information encoded in BERT embeddings has been utilized in parsing long sentences, but the performance enhancement is limited with the current architecture.

Error analysis shows that the TTS models are not simply applying a fixed Tone-3 sandhi pattern to all trisyllabic words. For different $[\sigma_3\sigma_3\sigma_3]_w$ trigrams, both TTS models are capable of varying their predictions either as $[\sigma_3\sigma_2\sigma_3]_w$ or $[\sigma_2\sigma_2\sigma_3]_w$ depending on the actual structure of the phrases, evidence that the TTS models have learned to generalize the rules over novel instances rather than simply memorizing a fixed pattern. But the low accuracy suggests that such generalizability is limited. In long sentences, most tonal errors arise from changing tones at incorrect syntactic boundaries rather than not changing tones at all. Both TTS models consistently applied Tone-3 sandhi rules to every bigram or trigram in a sentence, but the errors are likely to be caused by the failure to parse a sentence into its correct hierarchical structure. These results suggest that Tacotron 2 models, though capable of generating highly natural speech (including prosodic contours), still could not be used as a model for categorical Tone-3 sandhi in Mandarin.  

\begin{figure}[tbh]
    \centering
    \includegraphics[width=\linewidth]{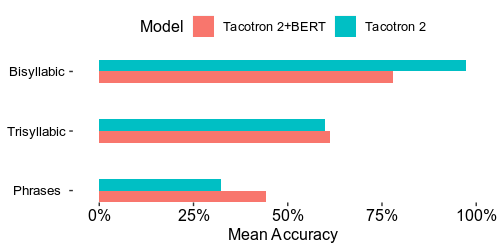}
    \caption{Mean accuracy of Tone-3 sandhi}
    \label{fig:acc}
\end{figure}

\section{Discussion}
In this study, linguistically informed stimuli have been used to probe the phonetic and phonological knowledge of TTS models. While both TTS models can produce the surface tonal coarticulations that closely match results from human speakers \cite{xu1997contextual}, both models still fail to correctly apply the Tone-3 sandhi rules to unseen complex stimuli. Supplying syntactic information through BERT embeddings helps but the overall accuracy of Tone-3 sandhi in phrases was still low, suggesting that the model does not fully acquire the underlying linguistic rules in the current data-driven approach, as shown in some similar NLP studies \cite{gulordava2018colorless,futrell2018rnns,futrell2019neural}. The rules induced by neural networks might not be as straightforward as those clear-cut algebraic rules summarized in linguistics \cite{baroni2019linguistic}. But this could also be caused by insufficient relevant instances in the training data, as the stimuli in this experiment are not highly frequent in daily conversations. These results also highlight the importance of developing linguistically informed stimuli to evaluate the performance of TTS models and of designing training data to cover as many phonological processes as possible.

In light of the results, TTS models seem to provide a complementary method to phonetic experiments for some research questions. The patterns of tonal coarticulation almost perfectly replicate what has been found in classic studies \cite{xu1997contextual}. Arguably TTS models can be applied to mine linguistic knowledge hidden in large quantities of coarsely annotated text-speech pairs, which are relatively easier to acquire than time-aligned phonemic transcriptions. Compared to the representation learning approach \cite{li2019representation,gerazov2018variational}, the use of TTS models directly addresses the problem of plausibility, as researchers can verify the speech samples through listening. Given the limitations of current TTS models, the research questions that can be explored in this approach are also limited, but nonetheless, with proper evaluation, they can be applied to generate novel hypotheses or provide additional evidence to experimental results.

\section{Acknowledgements}
I would like to express my gratitude to Professor Patrice Beddor and Professor Jelena Krivokapi\'{c} at University of Michigan for their helpful advice. I also thank my fellow linguists who had carefully listened to every generated sample in the experiment.

\bibliographystyle{IEEEtran}

\bibliography{template}

\begin{thebibliography}{10}
\providecommand{\url}[1]{#1}
\csname url@samestyle\endcsname
\providecommand{\newblock}{\relax}
\providecommand{\bibinfo}[2]{#2}
\providecommand{\BIBentrySTDinterwordspacing}{\spaceskip=0pt\relax}
\providecommand{\BIBentryALTinterwordstretchfactor}{4}
\providecommand{\BIBentryALTinterwordspacing}{\spaceskip=\fontdimen2\font plus
\BIBentryALTinterwordstretchfactor\fontdimen3\font minus
  \fontdimen4\font\relax}
\providecommand{\BIBforeignlanguage}[2]{{%
\expandafter\ifx\csname l@#1\endcsname\relax
\typeout{** WARNING: IEEEtran.bst: No hyphenation pattern has been}%
\typeout{** loaded for the language `#1'. Using the pattern for}%
\typeout{** the default language instead.}%
\else
\language=\csname l@#1\endcsname
\fi
#2}}
\providecommand{\BIBdecl}{\relax}
\BIBdecl

\bibitem{shen2018natural}
J.~Shen, R.~Pang, R.~J. Weiss, M.~Schuster, N.~Jaitly, Z.~Yang, Z.~Chen,
  Y.~Zhang, Y.~Wang, R.~Skerrv-Ryan \emph{et~al.}, ``Natural tts synthesis by
  conditioning wavenet on mel spectrogram predictions,'' in \emph{2018 IEEE
  International Conference on Acoustics, Speech and Signal Processing
  (ICASSP)}.\hskip 1em plus 0.5em minus 0.4em\relax IEEE, 2018, pp. 4779--4783.

\bibitem{li2018close}
N.~Li, S.~Liu, Y.~Liu, S.~Zhao, M.~Liu, and M.~Zhou, ``Close to human quality
  tts with transformer,'' \emph{arXiv preprint arXiv:1809.08895}, 2018.

\bibitem{ren2019fastspeech}
Y.~Ren, Y.~Ruan, X.~Tan, T.~Qin, S.~Zhao, Z.~Zhao, and T.-Y. Liu, ``Fastspeech:
  Fast, robust and controllable text to speech,'' \emph{arXiv preprint
  arXiv:1905.09263}, 2019.

\bibitem{oord2016wavenet}
A.~v.~d. Oord, S.~Dieleman, H.~Zen, K.~Simonyan, O.~Vinyals, A.~Graves,
  N.~Kalchbrenner, A.~Senior, and K.~Kavukcuoglu, ``Wavenet: A generative model
  for raw audio,'' \emph{arXiv preprint arXiv:1609.03499}, 2016.

\bibitem{mehri2016samplernn}
S.~Mehri, K.~Kumar, I.~Gulrajani, R.~Kumar, S.~Jain, J.~Sotelo, A.~Courville,
  and Y.~Bengio, ``Samplernn: An unconditional end-to-end neural audio
  generation model,'' \emph{arXiv preprint arXiv:1612.07837}, 2016.

\bibitem{prenger2019waveglow}
R.~Prenger, R.~Valle, and B.~Catanzaro, ``Waveglow: A flow-based generative
  network for speech synthesis,'' in \emph{ICASSP 2019-2019 IEEE International
  Conference on Acoustics, Speech and Signal Processing (ICASSP)}.\hskip 1em
  plus 0.5em minus 0.4em\relax IEEE, 2019, pp. 3617--3621.

\bibitem{futrell2018rnns}
R.~Futrell, E.~Wilcox, T.~Morita, and R.~Levy, ``Rnns as psycholinguistic
  subjects: Syntactic state and grammatical dependency,'' \emph{arXiv preprint
  arXiv:1809.01329}, 2018.

\bibitem{gulordava2018colorless}
K.~Gulordava, P.~Bojanowski, E.~Grave, T.~Linzen, and M.~Baroni, ``Colorless
  green recurrent networks dream hierarchically,'' \emph{arXiv preprint
  arXiv:1803.11138}, 2018.

\bibitem{prasad2019using}
G.~Prasad, M.~van Schijndel, and T.~Linzen, ``Using priming to uncover the
  organization of syntactic representations in neural language models,'' in
  \emph{Proceedings of the 23rd Conference on Computational Natural Language
  Learning (CoNLL)}, 2019, pp. 66--76.

\bibitem{linzen2019can}
T.~Linzen, ``What can linguistics and deep learning contribute to each other?
  response to pater,'' \emph{Language}, 2019.

\bibitem{li2019representation}
B.~Li, J.~Y. Xie, and F.~Rudzicz, ``Representation learning for discovering
  phonemic tone contours,'' \emph{arXiv preprint arXiv:1910.08987}, 2019.

\bibitem{gerazov2018variational}
B.~Gerazov, G.~Bailly, O.~Mohammed, Y.~Xu, and P.~N. Garner, ``A variational
  prosody model for mapping the context-sensitive variation of functional
  prosodic prototypes,'' \emph{arXiv preprint arXiv:1806.08685}, 2018.

\bibitem{chen2006production}
Y.~Chen and Y.~Xu, ``Production of weak elements in speech--evidence from f0
  patterns of neutral tone in standard chinese,'' \emph{Phonetica}, vol.~63,
  no.~1, pp. 47--75, 2006.

\bibitem{xu1997contextual}
Y.~Xu, ``Contextual tonal variations in mandarin,'' \emph{Journal of
  phonetics}, vol.~25, no.~1, pp. 61--83, 1997.

\bibitem{shen1990tonal}
X.~S. Shen, ``Tonal coarticulation in mandarin,'' \emph{Journal of Phonetics},
  vol.~18, no.~2, pp. 281--295, 1990.

\bibitem{chien2016priming}
Y.-F. Chien, J.~A. Sereno, and J.~Zhang, ``Priming the representation of
  mandarin tone 3 sandhi words,'' \emph{Language, Cognition and Neuroscience},
  vol.~31, no.~2, pp. 179--189, 2016.

\bibitem{shih1997mandarin}
C.~Shih, ``Mandarin third tone sandhi and prosodic structure,''
  \emph{Linguistic Models}, vol.~20, pp. 81--124, 1997.

\bibitem{fang2019towards}
W.~Fang, Y.-A. Chung, and J.~Glass, ``Towards transfer learning for end-to-end
  speech synthesis from deep pre-trained language models,'' \emph{arXiv
  preprint arXiv:1906.07307}, 2019.

\bibitem{hayashi2019pre}
T.~Hayashi, S.~Watanabe, T.~Toda, K.~Takeda, S.~Toshniwal, and K.~Livescu,
  ``Pre-trained text embeddings for enhanced text-to-speech synthesis,''
  \emph{Proc. Interspeech 2019}, pp. 4430--4434, 2019.

\bibitem{tyagi2019dynamic}
S.~Tyagi, M.~Nicolis, J.~Rohnke, T.~Drugman, and J.~Lorenzo-Trueba, ``Dynamic
  prosody generation for speech synthesis using linguistics-driven acoustic
  embedding selection,'' \emph{arXiv preprint arXiv:1912.00955}, 2019.

\bibitem{devlin2018bert}
J.~Devlin, M.-W. Chang, K.~Lee, and K.~Toutanova, ``Bert: Pre-training of deep
  bidirectional transformers for language understanding,'' \emph{arXiv preprint
  arXiv:1810.04805}, 2018.

\bibitem{cui2019pre}
Y.~Cui, W.~Che, T.~Liu, B.~Qin, Z.~Yang, S.~Wang, and G.~Hu, ``Pre-training
  with whole word masking for chinese bert,'' \emph{arXiv preprint
  arXiv:1906.08101}, 2019.

\bibitem{kingma2018glow}
D.~P. Kingma and P.~Dhariwal, ``Glow: Generative flow with invertible 1x1
  convolutions,'' in \emph{Advances in Neural Information Processing Systems},
  2018, pp. 10\,215--10\,224.

\bibitem{jadoul2018introducing}
Y.~Jadoul, B.~Thompson, and B.~De~Boer, ``Introducing parselmouth: A python
  interface to praat,'' \emph{Journal of Phonetics}, vol.~71, pp. 1--15, 2018.

\bibitem{rosenberg2017bias}
A.~Rosenberg and B.~Ramabhadran, ``Bias and statistical significance in
  evaluating speech synthesis with mean opinion scores.'' in
  \emph{Interspeech}, 2017, pp. 3976--3980.

\bibitem{futrell2019neural}
R.~Futrell, E.~Wilcox, T.~Morita, P.~Qian, M.~Ballesteros, and R.~Levy,
  ``Neural language models as psycholinguistic subjects: Representations of
  syntactic state,'' \emph{arXiv preprint arXiv:1903.03260}, 2019.

\bibitem{baroni2019linguistic}
M.~Baroni, ``Linguistic generalization and compositionality in modern
  artificial neural networks,'' \emph{arXiv preprint arXiv:1904.00157}, 2019.

\end{thebibliography}


\end{document}